\documentclass[runningheads]{llncs}
\usepackage{graphicx}

\usepackage{tikz}
\usepackage{comment}
\usepackage{amsmath,amssymb} 
\usepackage{color}
\usepackage{xcolor}
\usepackage{booktabs}
\usepackage{wrapfig}
\usepackage{caption}
\usepackage{subcaption}
\usepackage[boxed,ruled]{algorithm2e}

\newcommand{\keypoint}[1]{\vspace{0.1cm}\noindent\textbf{#1}\quad}
\newcommand{\diff}[1]{\scriptsize{#1}}
\newcommand{\wtpgd}[1]{WT-PGD}
\newcommand{\wtzoo}[1]{WT-ZOO}

\usepackage[accsupp]{axessibility}  

\DeclareMathOperator{\sign}{sign}

\begin{document}
\pagestyle{headings}
\mainmatter

\title{Attacking Adversarial Defences by Smoothing the Loss Landscape} 

\titlerunning{Attacking Adv. Defences by Smoothing the Loss Landscape}
%
\author{
Panagiotis Eustratiadis\inst{1} \and
Henry Gouk\inst{1} \and
Da Li\inst{1,2} \and
Timothy Hospedales\inst{1,2}
}
\authorrunning{P. Eustratiadis et al.}
%
\institute{
University of Edinburgh\\
\email{\{p.eustratiadis, henry.gouk, t.hospedales\}@ed.ac.uk}
\and
Samsung AI Center, Cambridge\\
\email{dali.academic@gmail.com}
}
\maketitle

\begin{abstract}
This paper investigates a family of methods for defending against adversarial attacks that owe part of their success to creating a noisy, discontinuous, or otherwise rugged loss landscape that adversaries find difficult to navigate. A common, but not universal, way to achieve this effect is via the use of stochastic neural networks.
We show that this is a form of gradient obfuscation, and propose a general extension to gradient-based adversaries based on the Weierstrass transform, which smooths the surface of the loss function and provides more reliable gradient estimates. We further show that the same principle can strengthen gradient-free adversaries.
We demonstrate the efficacy of our loss-smoothing method against both stochastic and non-stochastic adversarial defences that exhibit robustness due to this type of obfuscation. Furthermore, we provide analysis of how it interacts with Expectation over Transformation; a popular gradient-sampling method currently used to attack stochastic defences.
\end{abstract}

\section{Introduction}
\label{sec:introduction}

The discovery of adversarial examples in deep learning~\cite{iclr14intriguing}, together with its growing commercial and societal importance, has led to adversarial defence emerging as an important field of machine learning research, with the purpose of creating models that are robust against adversarial perturbations.
There is an interplay between adversarial attack and defence research, where stronger defences are developed, and often subsequently broken by more innovative attacks~\cite{kurakin2018adversarial}. An example of this dynamic is the discovery that many defences against gradient-based adversaries relied on  masking the gradient signal from the attacker~\cite{icml18obfuscation}. However, as shown by~\cite{icml18obfuscation}, such obfuscation gives a false sense of security and is easy to circumvent. They successfully attack stochastic defences by repeatedly sampling the gradient of the loss function w.r.t. the input and averaging the samples to obtain more reliable gradient estimates. They name this technique Expectation over Transformation (EoT)~\cite{icml18eot}. It has since been standardised that new stochastic defences~\cite{icml21wca,iccv19pni,cvpr20l2p,aaai2021sesnn} apply EoT during evaluation, to ensure that their apparent robustness does not rely on stochastic gradients.

In this paper, we reveal a form of gradient obfuscation that, to the best of our knowledge, is not yet known. So far, it is understood that stochastic neural networks (SNNs) defend effectively against adversarial attacks because having stochastic weights reduces overfitting, with similar effect to training the original neural network with Lipschitz regularisation~\cite{eccv18rse}, a property with strong theoretical links to adversarial robustness \cite{hein2017formal}. We show that there is an additional reason for their robust performance. Stochastic defences, even when averaging multiple gradient samples with EoT, tend to create a rough loss landscape that white-box adversaries find difficult to navigate. A second, and perhaps more interesting finding, is that this property is not exclusive to stochastic defences; there exist non-stochastic adversarial defences that have the same effect~\cite{icml21antiadv}.

We show that the aforementioned property can be attacked by an adversary. Specifically, we propose a stochastic extension to gradient-based attacks that approximates performing the Weierstrass Transform (WT)~\cite{bilodeau1962weierstrass,weierstrass1885analytische} on the loss function in order to smooth it before computing its gradient.
Interestingly, we find that the same method can be applied in a gradient-free setting to effectively circumvent the same type of obfuscation.

We experimentally support our insights by applying our extension to Projected Gradient Descent (PGD)~\cite{iclr18pgd} and other recent iterative FGSM variants \cite{iclr20nifgsm,cvpr21vmifgsm} as well as Zeroth Order Optimization (ZOO)~\cite{acm17zoo}, in the gradient-based and gradient-free settings respectively. We demonstrate the efficacy of our loss-smoothing method against both stochastic~\cite{icml21wca,iccv19pni,cvpr20l2p,aaai2021sesnn} and non-stochastic defences~\cite{iclr20kwta,icml21antiadv} that create a rough loss surface, and damage their robust performance by as much as 20\%.
Finally, we analyse how the WT interacts with EoT when attacking stochastic defences. We show that these two methods serve different purposes and are complementary. However, unlike an attack that applies EoT, a WT-based attack is effective against both stochastic and non-stochastic defences.

\section{Background and Related Work}
\label{sec:related-work}

We consider adversarial attacks under the $\ell_p$ threat model. For a clean input image $x$, an adversarial example $\tilde{x}$ is within the threat model if $||x - \tilde{x}||_p \leq \epsilon$, where $\epsilon$ is a small value indicating the attack strength, and $p$ is typically in $\{0, 2, \infty\}$.

\subsection{Gradient-Based Adversaries}
\label{sec:related-work-grad-adv}

Let $h_\theta$ be a classifier with parameters $\theta$, and $x$ an input image belonging to class $c \in C$.
The first and simplest gradient-based adversary outlined in prior work is the Fast Gradient Sign Method (FGSM)~\cite{iclr15fgsm}; a single-step attack that adds a small perturbation to $x$ in the direction indicated by the sign of the gradient of the loss function $\mathcal{L}(h_\theta(x), c)$ w.r.t. $x$. Formally,
\begin{equation}
  \tilde{x} = x + \epsilon \cdot \sign(\nabla_x\mathcal{L}(h_\theta(x), c))\;,
  \label{eq:fgsm}
\end{equation}
where $\epsilon$ denotes the attack strength.
The Basic Iterative Method (BIM)~\cite{iclr2017bim} was introduced shortly thereafter as an iterative variant of FGSM, followed by PGD~\cite{iclr18pgd}, an iterative variant of FGSM where the initial perturbation is a randomly selected point in the $\epsilon$-ball of $x$. Recent contributions have improved upon this scheme, e.g., through Nesterov's acceleration and variance tuning~\cite{iclr20nifgsm,cvpr21vmifgsm}.

It is currently common practice to use strong variants and extensions of PGD, such as PGD$^{100}$ and APGD~\cite{icml20auto} respectively, to evaluate newly-proposed adversarial defences~\cite{neurips21robustbench}. The \wtpgd; adversary proposed in this paper is also an extension of PGD.

\subsection{Dealing with Obfuscated Gradients}
\label{sec:related-work-eot}
In their paper,~\cite{icml18obfuscation} demonstrate that many existing defences create a false impression of robustness to gradient-based adversaries by masking the gradient of the loss function from the attacker. They identify three types of gradient obfuscation: shattered, stochastic, and vanishing gradients; and show that gradient-obfuscating defences are not reliable.

Stochastic gradients stem from defences where either the weights or the activations of SNNs are sampled from a distribution~\cite{eccv18rse,iclr19advbnn}. As a result, the gradient of their loss is also a distribution. To deal with stochastic gradients,~\cite{icml18obfuscation} applied EoT~\cite{icml18eot}, a method that repeatedly samples the target model's gradient w.r.t. the input, and computes the average of these samples to obtain the ``true'' gradient.
Following~\cite{icml18obfuscation}, it has become a requirement for stochastic defence research~\cite{icml21wca,iccv19pni,arxiv21graddiv} to incorporate a series of checks that ensure new stochastic defence methods do not owe their success to gradient obfuscation. Further, in order to circumvent non-stochastic, but otherwise obfuscating defences (e.g., shattered gradients), Gaussian sampling has been previously used~\cite{neurips20adaptive,icml21indicators}.

\keypoint{Expectation over Transformation}
We now highlight a few technical details about EoT. Let $h_\theta$ be a SNN with parameters $\theta$, and $x$ an input image belonging to class $c \in C$. The stochastic weights or activations of $h_\theta$ cause $h_\theta(x)$ to be randomised; as a result, $\nabla_{x}\mathcal{L}(h_\theta(x), c)$ is a distribution of gradients. EoT is, in essence, a Monte-Carlo sampling method that estimates the true gradient $\omega$ of the loss function by averaging $n$ gradient samples as
\begin{equation}
  \omega = \frac{1}{n}\sum_{i=0}^n\nabla_{x}\mathcal{L}(h_\theta^i(x), c)\;.
  \label{eq:eot}
\end{equation}

It is important to emphasise that the WT and EoT serve different purposes. Unlike our proposed method, detailed in Section~\ref{sec:method}, EoT has no ``spatial awareness'' of the loss' landscape, i.e., while applying EoT results in a better estimation of the gradient at $x$, it is uninformative regarding the gradient at $x+\delta$.
In this paper, we demonstrate that the WT and EoT are complementary, and maximally effective when used in combination.

\subsection{Defences with an Obfuscating Loss Landscape}

\begin{figure*}[t]
  \centering
  \begin{subfigure}{0.3\linewidth}
    \includegraphics[width=\textwidth]{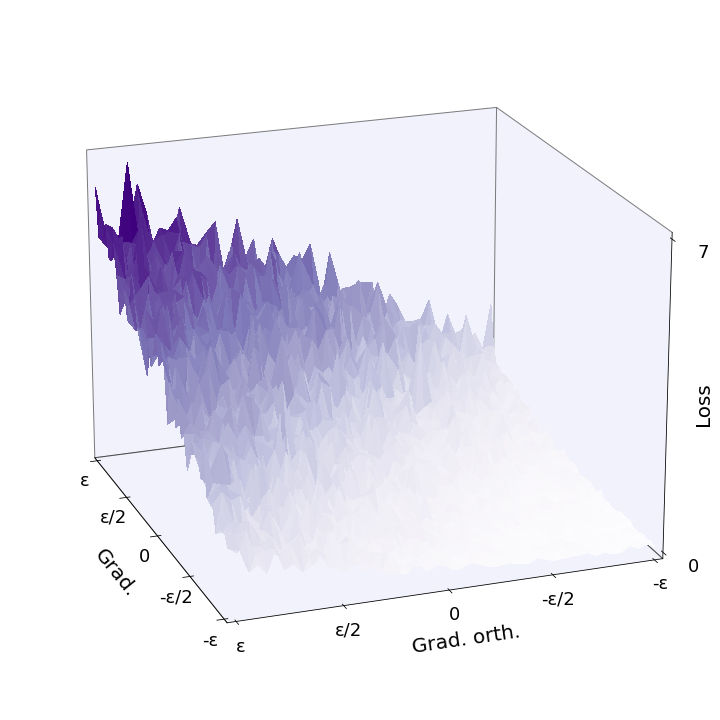}
    \caption{PNI~\cite{iccv19pni}}
    \label{fig:loss-landscapes-pni}
  \end{subfigure}
  \hfill
  \begin{subfigure}{0.3\linewidth}
    \includegraphics[width=\textwidth]{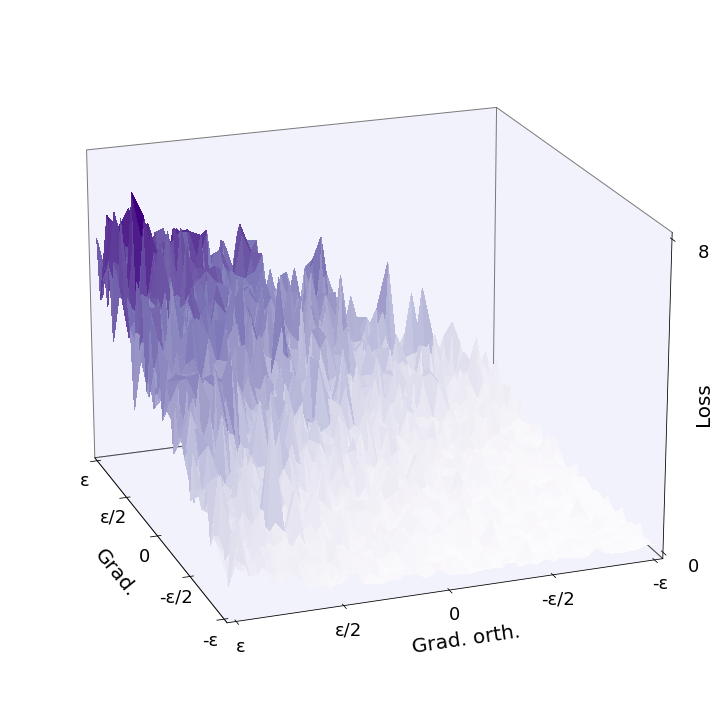}
    \caption{L2P~\cite{cvpr20l2p}}
    \label{fig:loss-landscapes-l2p}
  \end{subfigure}
  \hfill
  \begin{subfigure}{0.3\linewidth}
    \includegraphics[width=\textwidth]{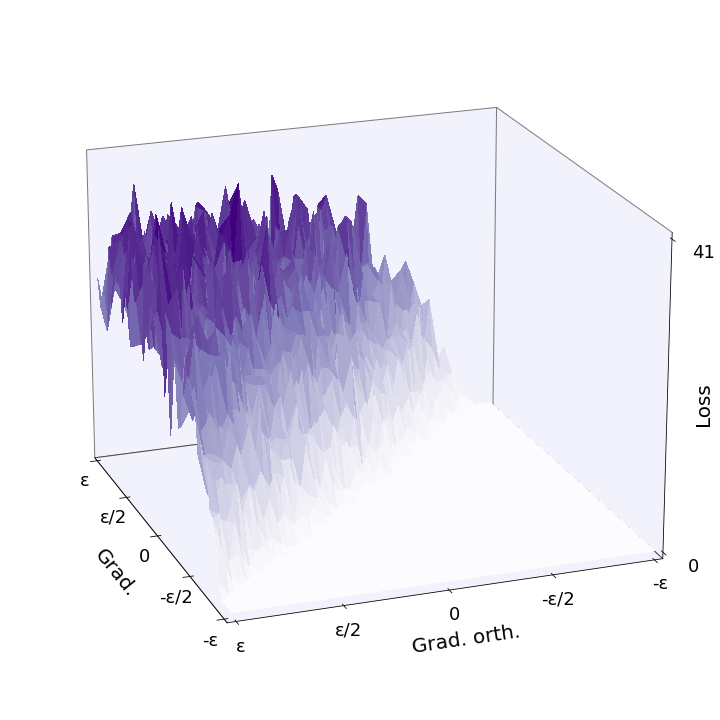}
    \caption{SE-SNN~\cite{aaai2021sesnn}}
    \label{fig:loss-landscapes-sesnn}
  \end{subfigure}
  \bigskip
  \begin{subfigure}{0.3\linewidth}
    \includegraphics[width=\textwidth]{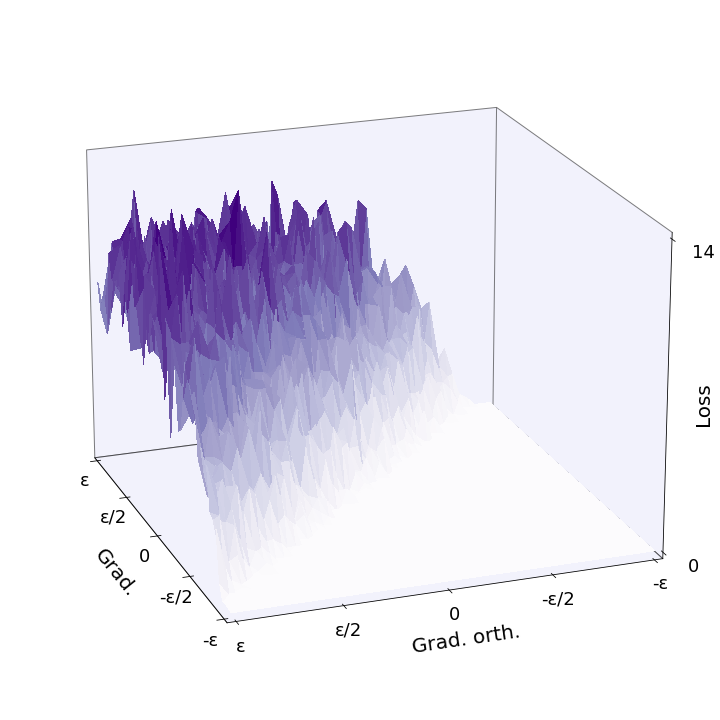}
    \caption{WCA~\cite{icml21wca}}
    \label{fig:loss-landscapes-wca}
  \end{subfigure}
  \hfill
  \begin{subfigure}{0.3\linewidth}
    \includegraphics[width=\textwidth]{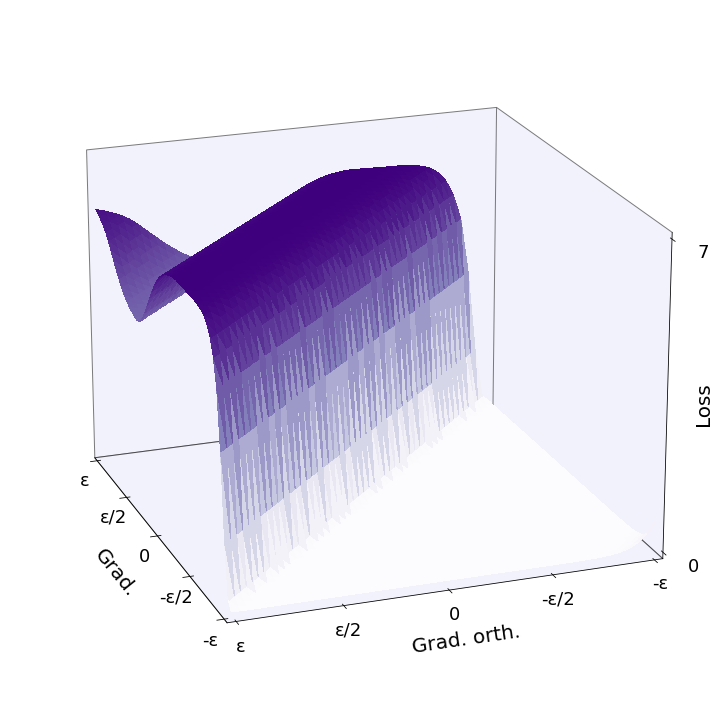}
    \caption{AA~\cite{icml21antiadv}}
    \label{fig:loss-landscapes-aa}
  \end{subfigure}
  \hfill
  \begin{subfigure}{0.3\linewidth}
    \includegraphics[width=\textwidth]{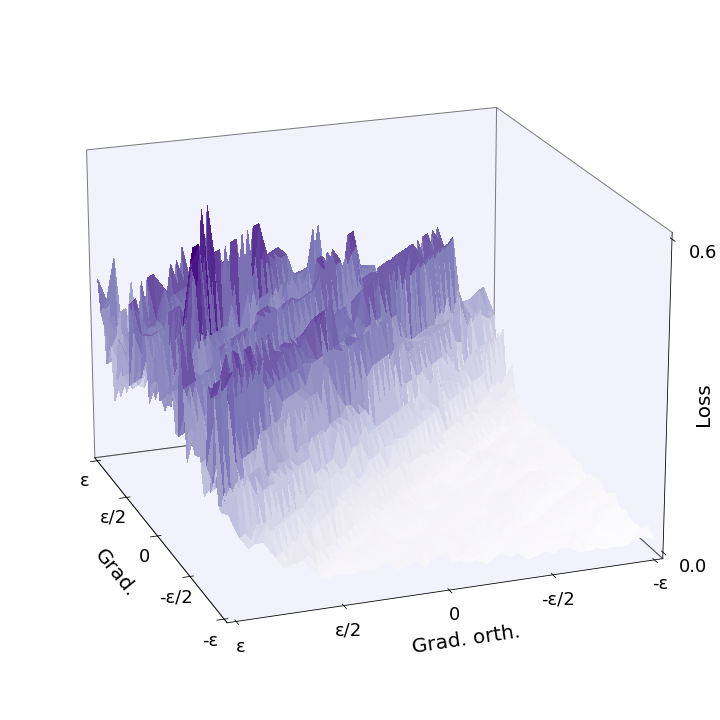}
    \caption{k-WTA~\cite{iclr20kwta}}
    \label{fig:loss-landscapes-kwta}
  \end{subfigure}
  \caption{Loss landscapes of each of the adversarial defences considered in this paper. All defences use a ResNet-18 backbone and the loss surfaces are constructed on a correctly-classified test image from CIFAR-10. The X axis is the gradient w.r.t. the clean input image, and the Y axis is chosen to be orthogonal to X. The Z axis is the value of the loss function for each perturbation within the $\epsilon$-ball of the input image, where $\epsilon = \frac{8}{255}$.}
  \label{fig:loss-landscapes}
\end{figure*}

We consider both stochastic and non-stochastic defences that we found to create a rough loss surface that is difficult for gradient-based adversaries to navigate. In the case of stochastic defences, we only consider related work that have applied EoT in their model evaluation.

Parametric Noise Injection (PNI)~\cite{iccv19pni} is a defence that equips convolutional neural network layers with additive noise drawn from an isotropic normal distribution.
Learn2Perturb (L2P)~\cite{cvpr20l2p} extends PNI to a richer noise model. Instead of learning a scalar intensity parameter $\alpha$, a noise injection module is learned that determines the strength of parameter-wise Gaussian noise injection at each layer.
Similarly to L2P, the Simple and Effective SNN (SE-SNN)~\cite{aaai2021sesnn}, learns a parameter-wise noise distribution motivated by the variational information bottleneck \cite{alemi2017vib}, and noise is only applied to the penultimate neural network layer.
Finally, Weight-Covariance Alignment (WCA)~\cite{icml21wca} extends the noise models above to include a full covariance (anisotropic) Gaussian noise model, thus generating correlated perturbations across channels. All the mentioned approaches \cite{icml21wca,iccv19pni,cvpr20l2p,aaai2021sesnn} include some noise-promoting regulariser to prevent the noise from shrinking to zero during training, with WCA's covariance alignment regulariser being derived from an adversarial generalisation bound in contrast to the prior models' heuristics. 

An obfuscating loss landscape is not an exclusive characteristic of SNNs. k-Winner Takes All (k-WTA)~\cite{iclr20kwta} is a defence that replaces the ReLU activation with a discontinuous function. Further, Anti-Adversaries (AA)~\cite{icml21antiadv} is a recent training-free adversarial defence that could be categorised as a ``black-box'' defence.
It improves adversarial robustness by prepending a layer that induces discontinuity to the loss landscape.

Our observation is that all these methods defend against white-box adversarial attacks largely through inducing rough loss landscapes that gradient-based adversaries struggle to ascend. Slices through the loss landscapes of the aforementioned defences are shown in Fig.~\ref{fig:loss-landscapes}, and we provide further details about how exactly they are computed in Appendix~\ref{appendix:compare}.

\section{Method}
\label{sec:method}

\subsection{The Weierstrass Transform}

The Weierstrass Transform (WT)~\cite{bilodeau1962weierstrass,weierstrass1885analytische} of a function $f$ is defined as the convolution of $f$ with a Gaussian kernel function $k$ in order to obtain $g$, a smoothed version of $f$. Formally,
\begin{equation}
  g(x) = \int_{-\infty}^{+\infty} k(x - y)\;f(y) \cdot dy,
  \quad k(x) = \frac{1}{\sqrt{4\pi}}\;e^{\frac{-x^2}{4}}\;.
  \label{eq:wt-analytical}
\end{equation}
The conventional Weierstrass Transform \cite{weierstrass1885analytische} is defined for functions of scalar variables and uses a Gaussian with a variance of $\sqrt{2}$. Because we are applying it to neural networks that are functions of many variables, and which may need to be smoothed to different extents, we relax these two conditions by using a multivariate Gaussian with a tuneable isotropic covariance matrix.

\subsection{Using the Weierstrass Transform to Attack}

Let $\mathcal{L}(h_\theta(x), c)$ be the classification loss function where $x$ is an input image belonging to a class $c \in C$, and $h_\theta$ a function approximator with parameters $\theta$. We can use Eq.~\ref{eq:wt-analytical} to define the smoothed loss function $\tilde{\mathcal{L}}$ as
\begin{equation}
  \tilde{\mathcal{L}}(h_\theta(x), c) = \int_{\mathbb{R}^d} k(x - y)\;\mathcal{L}(h_\theta(y), c) \cdot dy\;,
  \label{eq:wt-analytical-loss}
\end{equation}
where $d$ is the dimensionality of $x$. This can also be interpreted as an expectation
\begin{equation}
    \tilde{\mathcal{L}}(h_\theta(x), c) = \mathbb{E}_{\eta}[\mathcal{L}(h_\theta(x + \eta), c)],
    \quad \eta \sim \mathcal{N}(0, \sigma^2 I)\;.
\end{equation}

The dimensionality of the integral in  Eq.~\ref{eq:wt-analytical-loss} corresponds to the number of input pixels; so computing it directly is computationally infeasible. However, it is possible to compute a stochastic unbiased estimate of $\tilde{\mathcal{L}}$ by using Monte-Carlo sampling,
\begin{equation}
  \hat{\mathcal{L}}(h_\theta(x), c) = \frac{1}{m} \sum_{i=1}^m \mathcal{L}(h_\theta(X_i), c)\;,
  \label{eq:wt-sampling-loss}
\end{equation}
where $m$ is the number of perturbations sampled around $x$ and
\begin{equation}
  X_i = x + \eta_i,\quad \eta_i\sim\mathcal{N}(0, \sigma^2 I)\;.
  \label{eq:wt-sampling}
\end{equation}

The error introduced by this approximation of the WT is bounded (with high confidence), as shown in the following Theorem. It can be seen that the quality of the approximation improves as the number of samples, $m$, is increased.

\begin{theorem}
For a $k$-Lipschitz network, $h_\theta$, applied to a fixed instance $(x, c)$, and a loss function, $\mathcal{L}$, that is $L$-Lipschitz on the co-domain of $h_\theta$, we have with probability at least $1 - \delta$ that
\begin{equation}
    |\hat{\mathcal{L}}(h_\theta(x), c) - \tilde{\mathcal{L}}(h_\theta(x), c)| \leq kL\sigma\sqrt{\frac{4 d \textup{ln}(1/\delta)}{m}} + \frac{2kL\textup{ln}(1/\delta)}{3m}\;,
\end{equation}
where we assume that $x$ is contained within the unit ball in $d$-dimensional Euclidean space.
\end{theorem}
The proof of Theorem 1 is provided in Appendix~\ref{appendix:proof}.

\subsection{A Stochastic WT Extension of Gradient-Based Attacks}

\begin{algorithm}[t]
\DontPrintSemicolon
\SetKwInput{KwModel}{Model}
\KwData{$x$, $c$}
\KwModel{$h_\theta$}
\KwIn{$k$, $m$, $n$, $\alpha$, $\epsilon$, $\sigma$}
\KwOut{$\tilde{x}$}
$\tilde{x} \longleftarrow x + z,\quad z\sim\mathcal{U}(-\epsilon,\epsilon)$\;
\For{$k$ \upshape{iterations}} {
  $\tilde{X} \longleftarrow$ sample $m$ points around $\tilde{x}\;$ [Eq.~\ref{eq:wt-sampling}]\;
  \eIf{\upshape{defence is stochastic}} {
    $\omega \leftarrow \frac{1}{mn}\sum_{i=0}^m\sum_{j=0}^n\nabla_{x}\mathcal{L}(h_\theta^j(\tilde{X}_i), c)\;$ [Eq.~\ref{eq:wt-true-grad-with-eot}]\;
  } {
    $\omega \leftarrow \frac{1}{m}\sum_{i=0}^m\nabla_{x}\mathcal{L}(h_\theta(\tilde{X}_i), c)\;$ [Eq.~\ref{eq:wt-true-grad}]\;
  }
  $\tilde{x} \longleftarrow \tilde{x} + \alpha\,\sign(\omega) $\;
  project $\tilde{x}$ to $\ell_p$-ball of $\epsilon$\;
}
\caption{\wtpgd;}
\label{algo:wtpgd}
\end{algorithm}

Conceptually, any gradient-based adversary can be extended with the WT to smooth rugged loss landscapes and estimate the gradient of the loss more reliably. Algorithm~\ref{algo:wtpgd} describes \wtpgd;, our proposed method that is an extension of PGD. In addition to the standard hyperparameters of PGD, i.e., the number of iterations $k$, step size $\alpha$, and attack strength $\epsilon$, we add $m$ as the number of images sampled around $x$, and the standard deviation $\sigma$ of the zero-mean normal distribution from which the images are sampled.

\begin{figure}[t]
  \centering
  \begin{subfigure}{0.3\linewidth}
    \includegraphics[width=\textwidth]{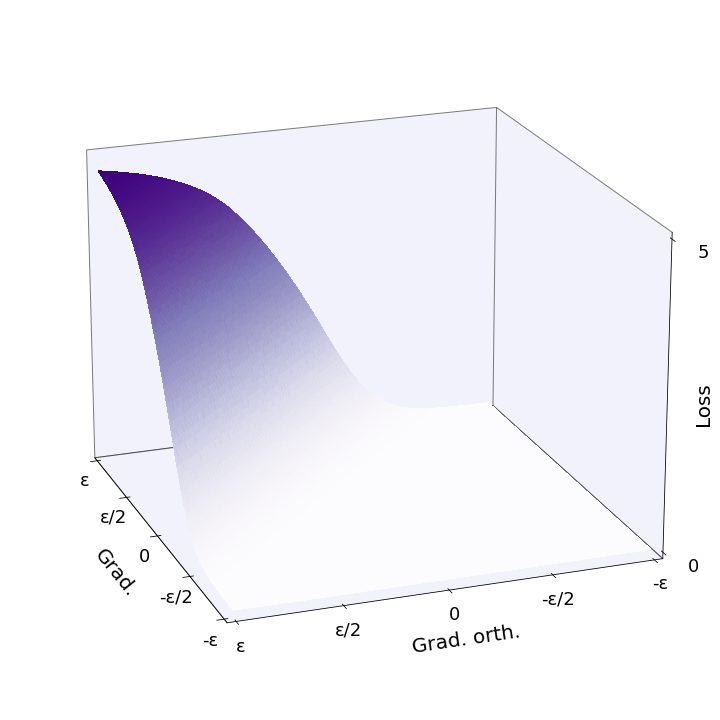}
    \caption{RN-18 (no defence).}
    \label{fig:loss-landscapes-resnet}
  \end{subfigure}
  \hfill
  \begin{subfigure}{0.3\linewidth}
    \includegraphics[width=\textwidth]{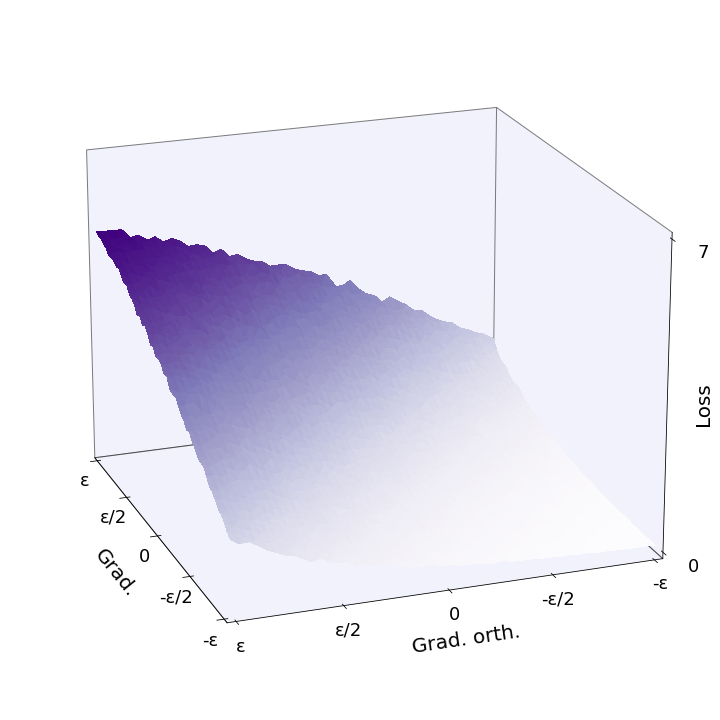}
    \caption{PNI + \wtpgd;}
    \label{fig:loss-landscapes-smoothed-pni}
  \end{subfigure}
  \hfill
  \begin{subfigure}{0.3\linewidth}
    \includegraphics[width=\textwidth]{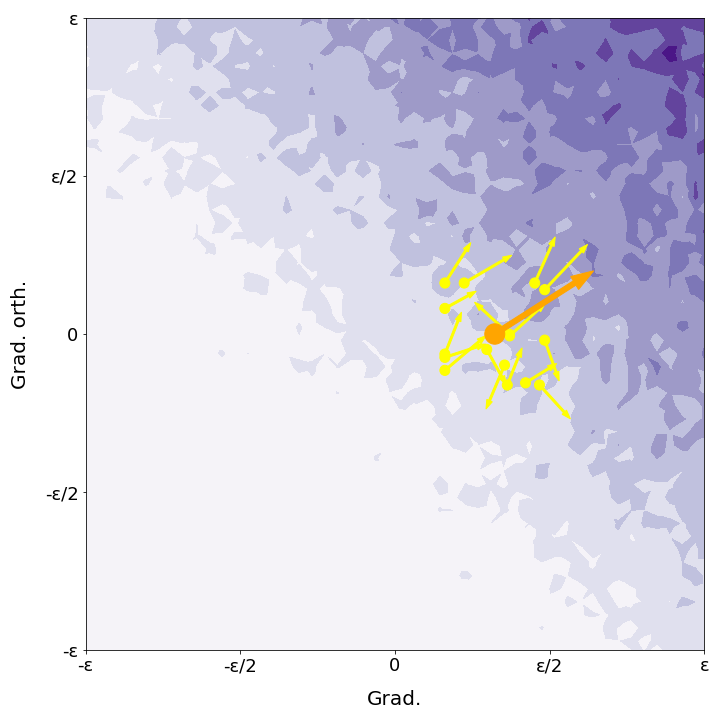}
    \caption{PNI (top-down)}
    \label{fig:wt-concept}
  \end{subfigure}
  \caption{Illustration of the intuition behind our WT attack. \textbf{Left:} Smooth surface of an undefended ResNet-18. \textbf{Middle:} When under attack by \wtpgd;, PNI's original noisy loss landscape (see Fig.~\ref{fig:loss-landscapes-pni}) is smoothed to better approximate one of an undefended network e.g., left figure. Refer to Fig.~\ref{fig:loss-landscapes-smoothed} in Appendix~\ref{appendix:compare} for the smoothed surfaces of all other defences. \textbf{Right:} Top-down view of Fig.~\ref{fig:loss-landscapes-pni}. The loss landscape around $x$ (dark orange point) is noisy and the adversary cannot find a reliable direction to follow. It therefore samples $m$ images around $x$ (yellow points) and follows the average gradient obtained at each of those points.}
  \label{fig:wt-concept-full}
\end{figure}

The main idea is that, given enough samples in close proximity to $x$, we can compute the true slope of the loss function as the average slope of the surface where these samples lie. Therefore, within the context of \wtpgd;, we define the true gradient $\omega$ as
\begin{equation}
  \omega = \frac{1}{m}\sum_{i=0}^m\nabla_{x}\mathcal{L}(h_\theta(\tilde{X}_i), c)\;,
  \label{eq:wt-true-grad}
\end{equation}
where $\tilde{X}$ denotes the set of images sampled around the perturbed image $\tilde{x}$, following Eq.~\ref{eq:wt-sampling}.

Fig.~\ref{fig:wt-concept} illustrates the concept of this attack. While the gradient at a particular image $x$ and samples nearby are individually noisy (random small yellow arrows), their aggregate direction (large orange arrow) ascends the loss surface.

\keypoint{Generalisation Properties}
Note that the WT only affects the gradient computation part of a gradient-based attack. In this paper we choose to illustrate the WT extension on PGD as a proof of concept, due to its convenient mathematical formulation as well as its efficacy as an attack. However, Eq.~\ref{eq:wt-true-grad} can effectively replace the gradient computation step in any gradient-based adversary \cite{iclr15fgsm,iclr20nifgsm,cvpr21vmifgsm}.

\subsubsection{Integration with EoT}
When we use Eq.~\ref{eq:wt-sampling-loss} and~\ref{eq:wt-sampling} to smooth the loss landscape of a stochastic defence, the gradient w.r.t. the input $x$, $\nabla_x\mathcal{L}(h_\theta(\tilde{X}), c)$, remains stochastic~\cite{icml18obfuscation}. It is therefore sensible to apply EoT~\cite{icml18eot} on the sampled $\tilde{X}$, and average over the output distribution of $h_\theta$. Incorporating Eq.~\ref{eq:eot} into Eq.~\ref{eq:wt-true-grad} we get
\begin{equation}
  \omega = \frac{1}{mn}\sum_{i=0}^m\sum_{j=0}^n\nabla_{x}\mathcal{L}(h_\theta^j(\tilde{X}_i), c)\;.
  \label{eq:wt-true-grad-with-eot}
\end{equation}
A thorough empirical analysis of how the WT interacts with EoT is presented in Section~\ref{sec:wt-vs-eot}, along with an ablation study for each individual component.

\subsection{A Stochastic WT Extension of Gradient-Free Attacks}

Although we primarily focus on the WT as an extension of gradient-based attacks, its potential impact when applied to gradient-free attacks cannot be ignored. In Appendix~\ref{appendix:gradient-free} we demonstrate WT's generality by integrating it with ZOO~\cite{acm17zoo}, a black-box adversary that uses gradient approximation instead of surrogate models~\cite{acm17zoo,arxiv16transfer,asiaccs17blackbox}, assuming access only to the per-class posterior $p\big(h(x)\big)$.

\begin{table}[t]
  \centering
  \caption{Robust accuracy \% of PGD and \wtpgd; attacks on CIFAR. All defences use a RN-18 backbone.}
  \resizebox{1.\linewidth}{!}{
  \begin{tabular}{@{}lcccccccc@{}}
    \toprule
    & \multicolumn{4}{c}{CIFAR-10} & \multicolumn{4}{c}{CIFAR-100} \\
    \cmidrule(lr){2-5}\cmidrule(lr){6-9}
    Method & PGD$_{10}$ & \wtpgd;$_{10}$ & PGD$_{100}$ & \wtpgd;$_{100}$
           & PGD$_{10}$ & \wtpgd;$_{10}$ & PGD$_{100}$ & \wtpgd;$_{100}$ \\
    \midrule
    PNI    & 49.4 & 34.8 \diff{(-14.6)}
           & 31.4 & 13.7 \diff{(-17.7)}
           & 22.2 & 17.9 \diff{(\hphantom{ }-4.3)}
           & 10.1 & \hphantom{0}9.4 \diff{(\hphantom{ }-0.7)} \\
    L2P    & 56.1 & 47.2 \diff{(\hphantom{ }-8.9)}
           & 20.5 & 18.2 \diff{(\hphantom{ }-2.3)}
           & 26.1 & 11.5 \diff{(-14.6)}
           & 18.4 & 10.3 \diff{(\hphantom{ }-8.1)} \\
    SE-SNN & 39.8 & 21.3 \diff{(-18.5)}
           & 13.9 & 12.5 \diff{(\hphantom{ }-1.4)}
           & 18.6 & \hphantom{0}8.0 \diff{(-10.6)}
           & 15.9 & \hphantom{0}5.9 \diff{(-10.0)} \\
    WCA    & 61.7 & 53.3 \diff{(\hphantom{ }-8.4)}
           & 58.6 & 37.6 \diff{(-21.0)}
           & 41.7 & 27.4 \diff{(-14.3)}
           & 39.0 & 10.8 \diff{(-28.2)} \\
    AA     & 63.2 & 43.9 \diff{(-19.3)}
           & 43.6 & 25.9 \diff{(-17.7)}
           & 47.9 & 29.6 \diff{(-18.3)}
           & 43.6 & 21.2 \diff{(-22.4)} \\
    k-WTA  & 58.0 & 33.1 \diff{(-24.9)}
           & 48.2 & 30.7 \diff{(-17.5)}
           & 44.3 & 24.1 \diff{(-20.2)}
           & 37.5 & 15.3 \diff{(-22.2)} \\
    \bottomrule
  \end{tabular}
  }
  \label{tab:eval-cifar}
\end{table}
\begin{table}[t]
  \centering
  \caption{Robust accuracy \% of PGD and \wtpgd; attacks on CIFAR-100 and Imagenette (full-resolution). All defences use a WRN-34-10 backbone.}
  \resizebox{1.\linewidth}{!}{
  \begin{tabular}{@{}lcccccccc@{}}
    \toprule
    & \multicolumn{4}{c}{CIFAR-100} & \multicolumn{4}{c}{Imagenette} \\
    \cmidrule(lr){2-5}\cmidrule(lr){6-9}
    Method & PGD$_{10}$ & \wtpgd;$_{10}$ & PGD$_{100}$ & \wtpgd;$_{100}$
           & PGD$_{10}$ & \wtpgd;$_{10}$ & PGD$_{100}$ & \wtpgd;$_{100}$ \\
    \midrule
    PNI    & 51.6 & 32.5 \diff{(-19.1)}
           & 48.4 & 31.3 \diff{(-17.1)}
           & 51.8 & 39.6 \diff{(-12.2)}
           & 42.3 & 24.3 \diff{(-18.0)} \\
    L2P    & 45.3 & 32.4 \diff{(-12.9)}
           & 40.0 & 29.5 \diff{(-10.5)}
           & 63.4 & 46.9 \diff{(-16.5)}
           & 42.4 & 23.2 \diff{(-19.2)} \\
    SE-SNN & 44.6 & 34.9 \diff{(\hphantom{ }-9.7)}
           & 46.0 & 31.0 \diff{(-15.0)}
           & 47.2 & 22.9 \diff{(-24.3)}
           & 41.1 & 21.7 \diff{(-19.4)} \\
    WCA    & 63.6 & 54.5 \diff{(\hphantom{ }-9.1)}
           & 56.7 & 44.5 \diff{(-12.2)}
           & 67.5 & 51.0 \diff{(-16.5)}
           & 50.3 & 35.6 \diff{(-14.7)} \\
    AA     & 76.1 & 59.2 \diff{(-16.9)}
           & 62.4 & 54.0 \diff{(\hphantom{ }-8.4)}
           & 69.3 & 44.8 \diff{(-24.5)}
           & 57.1 & 39.4 \diff{(-17.7)} \\
    k-WTA  & 60.2 & 46.1 \diff{(-14.1)}
           & 51.3 & 34.4 \diff{(-16.9)}
           & 55.7 & 33.6 \diff{(-22.1)}
           & 52.0 & 28.3 \diff{(-23.7)} \\
    \bottomrule
  \end{tabular}
  }
  \label{tab:eval-wrn34-10}
\end{table}
\begin{table}[t]
  \centering
  \caption{Robust accuracy \% of SI-NI-FGSM (F1, \cite{iclr20nifgsm}) and VMI-FGSM (F2, \cite{cvpr21vmifgsm}) attacks and their respective WT extensions on CIFAR (RN-18 backbone) and Imagenette (WRN-34-10 backbone). Names are shortened for better readability.}
  \resizebox{1.\linewidth}{!}{
  \begin{tabular}{@{}lcccccccccccc@{}}
    \toprule
    & \multicolumn{4}{c}{CIFAR-10} & \multicolumn{4}{c}{CIFAR-100} & \multicolumn{4}{c}{Imagenette} \\
    \cmidrule(lr){2-5}\cmidrule(lr){6-9}\cmidrule(lr){10-13}
    Method & (F1) & WT-(F1) & F2 & WT-(F2)
           & (F1) & WT-(F1) & F2 & WT-(F2)
           & (F1) & WT-(F1) & F2 & WT-(F2) \\
    \midrule
    PNI    & 48.2 & 35.5 \diff{(-12.7)}
           & 38.3 & 27.4 \diff{(-10.9)}
           & 24.9 & 13.0 \diff{(-11.9)}
           & 25.7 & 18.6 \diff{(\hphantom{ }-7.1)}
           & 47.4 & 37.2 \diff{(-10.2)}
           & 42.5 & 33.2 \diff{(\hphantom{ }-9.3)} \\
    L2P    & 56.1 & 44.9 \diff{(-11.2)}
           & 31.7 & 19.2 \diff{(-12.5)}
           & 27.2 & 18.5 \diff{(\hphantom{ }-8.7)}
           & 30.1 & 21.0 \diff{(\hphantom{ }-9.1)}
           & 59.6 & 46.1 \diff{(-13.5)}
           & 42.4 & 30.5 \diff{(-11.9)} \\
    SE-SNN & 40.5 & 31.6 \diff{(\hphantom{ }-8.9)}
           & 38.1 & 22.8 \diff{(-15.3)}
           & 25.3 & 12.2 \diff{(-13.1)}
           & 28.9 & 15.0 \diff{(-13.9)}
           & 44.8 & 33.9 \diff{(-10.9)}
           & 40.7 & 38.4 \diff{(\hphantom{ }-2.3)} \\
    WCA    & 58.5 & 54.0 \diff{(\hphantom{ }-4.5)}
           & 55.7 & 34.8 \diff{(-20.9)}
           & 45.8 & 30.4 \diff{(-15.4)}
           & 44.0 & 33.2 \diff{(-10.8)}
           & 64.0 & 59.0 \diff{(\hphantom{ }-5.0)}
           & 51.6 & 42.3 \diff{(\hphantom{ }-9.3)} \\
    AA     & 61.8 & 53.6 \diff{(\hphantom{ }-8.2)}
           & 58.0 & 41.4 \diff{(-16.6)}
           & 46.7 & 31.8 \diff{(-14.9)}
           & 41.1 & 23.3 \diff{(-17.8)}
           & 66.5 & 49.3 \diff{(-17.2)}
           & 56.9 & 43.0 \diff{(-13.9)} \\
    k-WTA  & 55.3 & 43.0 \diff{(-12.3)}
           & 46.9 & 38.9 \diff{(\hphantom{ }-8.0)}
           & 49.4 & 38.0 \diff{(-11.4)}
           & 37.2 & 27.6 \diff{(\hphantom{ }-9.6)}
           & 57.9 & 46.5 \diff{(-11.4)}
           & 46.6 & 38.7 \diff{(\hphantom{ }-7.9)} \\
    \bottomrule
  \end{tabular}
  }
  \label{tab:eval-fgsm-variants}
\end{table}
\begin{table}[t]
  \centering
  \caption{Robust accuracy scores \% of gradient-free attacks ZOO and \wtzoo; on CIFAR (RN-18 backbone) and Imagenette (WRN-34-10 backbone).}
  \resizebox{.7\linewidth}{!}{
  \begin{tabular}{@{}lcccccc@{}}
    \toprule
    & \multicolumn{2}{c}{CIFAR-10} & \multicolumn{2}{c}{CIFAR-100} & \multicolumn{2}{c}{Imagenette} \\
    \cmidrule(lr){2-3}\cmidrule(lr){4-5}\cmidrule(lr){6-7}
    Method & ZOO & WT-ZOO
           & ZOO & WT-ZOO
           & ZOO & WT-ZOO \\
    \midrule
    PNI    & 62.1 & 54.3 \diff{(\hphantom{ }-7.8)}
           & 38.1 & 25.7 \diff{(-12.4)}
           & 59.2 & 41.0 \diff{(-18.2)} \\
    L2P    & 63.7 & 56.1 \diff{(\hphantom{ }-7.6)}
           & 37.5 & 29.7 \diff{(\hphantom{ }-7.8)}
           & 65.8 & 54.3 \diff{(-11.5)} \\
    SE-SNN & 59.4 & 44.3 \diff{(-15.1)}
           & 28.3 & 21.5 \diff{(\hphantom{ }-6.8)}
           & 49.8 & 37.6 \diff{(-12.2)} \\
    WCA    & 70.9 & 64.8 \diff{(\hphantom{ }-6.1)}
           & 48.8 & 42.8 \diff{(\hphantom{ }-6.0)}
           & 72.3 & 61.9 \diff{(-10.4)} \\
    AA     & 74.1 & 66.5 \diff{(\hphantom{ }-7.6)}
           & 52.7 & 42.3 \diff{(-10.4)}
           & 77.9 & 60.6 \diff{(-17.3)} \\
    k-WTA  & 70.2 & 64.5 \diff{(-5.7)}
           & 55.2 & 43.2 \diff{(-12.0)}
           & 70.1 & 53.7 \diff{(-16.4)} \\
    \bottomrule
  \end{tabular}
  }
  \label{tab:eval-zoo}
\end{table}

\section{Experiments}
\label{sec:experiments}

\subsection{Experimental Setup}
For our experiments we consider four stochastic defences (PNI~\cite{iccv19pni}, L2P~\cite{cvpr20l2p}, SE-SNN~\cite{aaai2021sesnn} and WCA~\cite{icml21wca}) and two non-stochastic (k-WTA~\cite{iclr20kwta} and AA~\cite{icml21antiadv}). For fair comparison these defences use the same backbone architecture, ResNet-18 (RN-18) and Wide ResNet-34-10 (WRN-34-10)~\cite{cvpr16resnet,bmvc16wideresnet} in the corresponding experiments.
We evaluate their performance against the gradient-based \wtpgd;$_{10}$ and \wtpgd;$_{100}$, and the gradient-free \wtzoo;. In terms of datasets, we consider CIFAR-10, CIFAR-100~\cite{cifarCitation} and Imagenette~\cite{imagenette} with high-resolution images.
Our hyperparameter selection is outlined in Appendix~\ref{appendix:hyperparameters}.

\subsection{Quantitative Evaluation}
\label{sec:evaluation}

In Tables~\ref{tab:eval-cifar} and~\ref{tab:eval-wrn34-10} we report the accuracy of our selection of adversarial defences when under our \wtpgd; attack against the baselines. It is evident that \wtpgd; outperforms PGD consistently across defences, benchmarks, for different attack strength and network depth.
In particular, we can see that: (i) Every defence considered suffers substantially; in some cases even with more than -20\% in robust accuracy. (ii) Weaker defences are broken near completely, with L2P, SE-SNN, and k-WTA failing on CIFAR-10; and PNI, L2P, SE-SNN and k-WTA failing on CIFAR-100. (iii) The stronger WCA and AA defences tend to suffer large hits, especially under \wtpgd;$_{100}$. (iv) Our attack is particularly effective with high-resolution images, with most defenses suffering a performance reduction of over 15\%.

To show the generality of our method, we apply the WT extension to the more sophisticated and recently proposed gradient-based adversaries NI-FGSM~\cite{iclr20nifgsm} and VMI-FGSM~\cite{cvpr21vmifgsm} that use Nesterov's acceleration and variance tuning to improve attack strength and transferability.
Table~\ref{tab:eval-fgsm-variants} shows results consistent with our previous evaluation, and proves that our loss-smoothing method can effectively strengthen recently proposed attacks of higher complexity than PGD.
Finally, in Table~\ref{tab:eval-zoo} we present our evaluation of \wtzoo;. It is evident that even though (i) the performance reduction is on average slightly lower than the gradient-based setting and (ii) WT-ZOO imposes an additional query-efficiency cost, WT-ZOO is still successful in attacking these obfuscating defences. 

These experimental results support that rugged loss surfaces can be exploited, and loss-smoothing adversaries are significantly stronger against this type of gradient obfuscation.

\subsection{Interaction between WT and EoT}
\label{sec:wt-vs-eot}

\begin{table}[t]
  \centering
  \caption{Ablation study: effect of the WT and EoT individually against stochastic defences. The scores are the robust accuracy \% on CIFAR-10.}
  \resizebox{\linewidth}{!}{
  \begin{tabular}{@{}lcccc@{}}
    \toprule
    (Attack: \wtpgd;$_{10}$)& No WT + No EoT & No WT + EoT$_{16}$ & WT$_{16}$ + No EoT & WT$_{16}$ + EoT$_{16}$ \\
    \midrule
    PNI    & 50.6 & 49.1 & 48.7 & 34.8 \\
    L2P    & 58.9 & 54.4 & 55.0 & 47.2 \\
    SE-SNN & 46.6 & 39.5 & 39.7 & 21.3 \\
    WCA    & 72.0 & 58.4 & 61.1 & 53.3 \\
    \bottomrule
  \end{tabular}
  }
  \label{tab:ablation-study}
\end{table}

In this Section we analyse how the WT interacts with EoT when attacking stochastic defences. An ablation study is presented in Table~\ref{tab:ablation-study}, where we evaluate the two methods individually and in combination when attacking PNI, L2P, SE-SNN and WCA. We start by setting the baseline to regular PGD, and then vary each of the two components by setting the number of WT samples and EoT iterations to 16 (Appendix~\ref{appendix:ablation} explains why 16), to keep consistent with our evaluation in Section~\ref{sec:evaluation}.

Our ablation study shows that, while each method increases attack strength, neither is significantly better than the other in terms of individual performance. We conclude the WT and EoT are most effective when used in combination, to deal with the noisy loss landscape and the stochastic gradients respectively. Further analysis on this is provided in Appendix~\ref{appendix:ablation}.

\section{Conclusions}
\label{sec:conclusions}

We reveal a new form of gradient obfuscation that can be a property of stochastic, as well as non-stochastic adversarial defences. This gradient obfuscation occurs when a defence creates a noisy or discontinuous loss landscape to mislead gradient-based adversaries. This does not constitute an adequate defence, and can be circumvented by smoothing the surface of the loss function before following the gradient w.r.t. the input. We propose a smoothing method with which both gradient-based and gradient-free adversaries can be extended, utilising a Monte-Carlo variant of the Weierstrass transform.
As demonstrated by applying the WT on PGD, ZOO and [SI-NI/VMI]-FGSM, this extension enables strong, successful attacks.

We further illustrate the smoothing capabilities of our adversary beyond the quantitative evaluation presented in Section~\ref{sec:evaluation}, by plotting the loss surfaces of the defences before and after WT smoothing (Fig.~\ref{fig:loss-landscapes} main paper and Fig.~\ref{fig:loss-landscapes-smoothed} Appendix~\ref{appendix:compare}).
We hope that highlighting this novel type of attack against this class of adversarial defences will inspire future research to avoid relying on this weak defence strategy for robustness.

%
%
\bibliographystyle{splncs04}
\bibliography{main}

\newpage
\appendix
\onecolumn

\section{Source Code}

The source code for (i) \wtpgd; and \wtzoo; and (ii) our diagnostic tool for visualising a neural network's loss landscape is publicly available on GitHub\footnote{\url{https://github.com/peustr/wt-pgd}}.

\section{Proof of Theorem 1}
\label{appendix:proof}

\begin{proof}
The proof is based on using a Bernstein inequality. Let $Z_1$, ..., $Z_m$ be independent random variables taking positive values in $[a, b]$, and let $S = \frac{1}{m}\sum_{i}^m Z_i$. From \cite{lafferty2010concentration}, Bernstein's inequality tells that
\begin{equation}
    P(|S - \mathbb{E}[S]| > t) \leq 2 \textup{exp} \Bigg ( \frac{-mt^2}{2 \textup{Var}[S] + \frac{2}{3}rt} \Bigg )\;,
\end{equation}
where $r = b - a$. By setting $\delta = P(|S - \mathbb{E}[S]| > t)$ this can be rearranged to show that, with probability at least $1 - \delta$,
\begin{equation}
    |S - \mathbb{E}[S]| \leq \sqrt{\frac{2 \textup{Var}[S] \textup{ln}(1/\delta)}{m}} + \frac{2r\textup{ln}(1/\delta)}{3m}\;.
\end{equation}
The result follows from using $Z_i = \mathcal{L}(h_\theta(X_i), c)$ and upper bounding $\textup{Var}[S]$ and $r$. Because $h_\theta$ is $k$-Lipschitz and $\mathcal{L}$ is $L$-Lipschitz on the co-domain of $h_\theta$, we can say that $\mathcal{L}(h_\theta(\cdot), \cdot)$ is $kL$-Lipschitz. From this Lipschitz property, we know that $b \leq a + kL$, and therefore $r \leq kL$.

Denote by $X_i^\prime$ and $S^\prime$ random variables that follow the same distribution as $X_i$ and $S$, respectively. The bound for the variance arises from
\begin{align}
    &\textup{Var}[S] \\
    &= \mathbb{E}_S[(\mathbb{E}_{S^\prime}[S^\prime] - S)^2] \\
    &\leq \mathbb{E}_{X_i}\mathbb{E}_{X_i^\prime} \Bigg [ \Big ( \frac{1}{m} \sum_{i=1}^m (\mathcal{L}(h_\theta(X_i^\prime), c) - \mathcal{L}(h_\theta(X_i), c)) \Big )^2 \Bigg ] \\
    &\leq \mathbb{E}_{X_i} \mathbb{E}_{X_i^\prime} \Big [ \|X_i^\prime - X_i\|_2^2 k^2L^2 \Big ] \\
    &= 2k^2L^2d\sigma^2,
\end{align}
where the first inequality is due to Jensen's inequality, and the second is from the Lipschitz property of the model. The final equality arises because $X^\prime - X \sim \mathcal{N}(0, 2\sigma^2I)$, and the expected value of the squared Euclidean norm of a sample from a Gaussian distribution is the trace of the covariance matrix.
\end{proof}

\section{A Stochastic WT Extension of Gradient-Free Attacks}
\label{appendix:gradient-free}

\begin{algorithm}[t]
\DontPrintSemicolon
\SetKwInput{KwModel}{Model}
\KwData{$x^d$, $c$}
\KwModel{$h$}
\KwIn{$k$, $m$, $n$, $\alpha$, $\epsilon$, $\sigma$}
\KwOut{$\tilde{x}$}
\For{$k$ \upshape{iterations}} {
  Randomly pick coordinates $\vec{\rho}\in\{1, \dots, d\}$\;
  $\tilde{X} \longleftarrow$ sample $m$ points around $\tilde{x}\;$ [Eq.~\ref{eq:wt-sampling}]\;
  \eIf{\upshape{defence is stochastic}} {
    $\delta^* \leftarrow \frac{1}{mn}\sum_{i=0}^m\sum_{j=0}^n \delta_j(X_i, c)\;$[Eq.~\ref{eq:wt-true-grad-with-eot-zoo}]\;
  } {
    $\delta^* \leftarrow \frac{1}{m}\sum_{i=0}^m \delta(X_i, c)\;$[Eq.~\ref{eq:wt-true-grad-zoo}]\;
  }
  $\tilde{x} \longleftarrow \tilde{x} + \delta^*$\;
  project $\tilde{x}$ to $\ell_p$-ball of $\epsilon$\;
}
\caption{\wtzoo; (Newton's Coordinate Descent)}
\label{algo:wtzoo}
\end{algorithm}

Given an input image $x$ and a pixel coordinate $\rho$, ZOO iteratively constructs a perturbation $\delta$ on $x_\rho$ as
\begin{equation}
    \delta(x, c) =
    \begin{cases}
        -\alpha \hat{g}_\rho(x, c) &\text{$\hat{h}_\rho \leq 0$}
        \\
        -\alpha \frac{\hat{g}_\rho(x, c)}{\hat{h}_\rho(x, c)} &\text{otherwise}
    \end{cases}\;,
    \label{eq:zoo-delta}
\end{equation}
where $\alpha$ denotes the learning rate. $\hat{g}_i$ and $\hat{h}_i$ are the first- and second-order approximate gradients of a hinge-like loss function
\begin{equation}
    f(x,c_0) = \max\{\log h(x)_{c_0} - \max_{c \neq c_0}\log h(x)_c, -\kappa\}\;,
\end{equation}
where $\kappa \geq 0$. Algorithm~\ref{algo:wtzoo} details \wtzoo;. Note that the principle behind the WT extension remains the same as in the white-box setting. Adapting Eq.~\ref{eq:wt-true-grad} and~\ref{eq:wt-true-grad-with-eot} with ZOO's gradient approximation (Eq.~\ref{eq:zoo-delta}) we respectively get
\begin{equation}
    \delta^* = \frac{1}{m}\sum_{i=0}^m \delta(X_i, c)\;,
    \label{eq:wt-true-grad-zoo}
\end{equation}
and for stochastic defences
\begin{equation}
    \delta^* = \frac{1}{mn}\sum_{i=0}^m\sum_{j=0}^n \delta_j(X_i, c)\;.
  \label{eq:wt-true-grad-with-eot-zoo}
\end{equation}
As ZOO estimates gradients with finite difference it is susceptible to being mislead by a rough loss surface (Fig.~\ref{fig:loss-landscapes}). Smoothing the loss estimates at each point improves the quality of approximate gradient estimation for the ZOO attacker.

\section{Experimental Setup: Hyperparameters}
\label{appendix:hyperparameters}

\begin{figure}[t]
  \centering
  \includegraphics[width=0.35\linewidth]{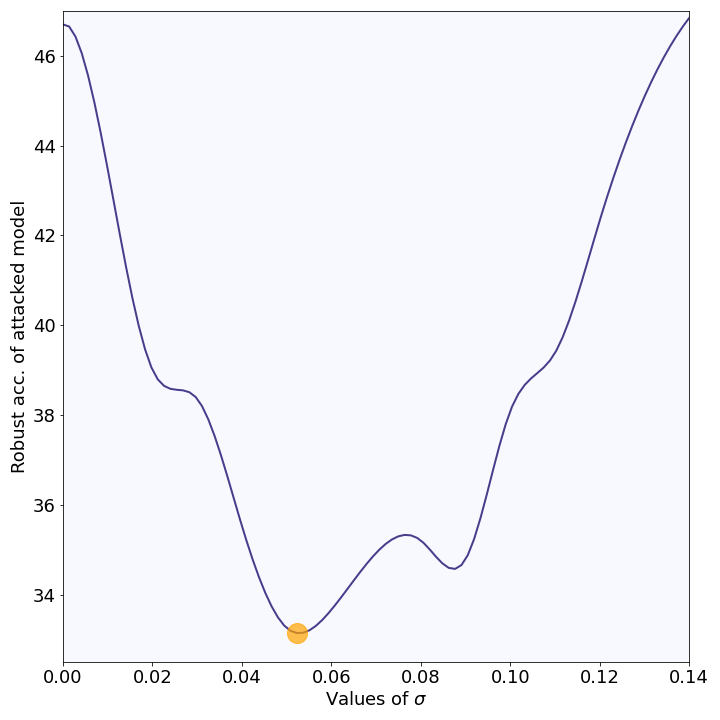}
  \caption{Sensitivity study of $\sigma$. If the value of $\sigma$ is either too low or too high, the attack is not as effective. The local minima in this curve are caused by randomness and are slightly different in each execution, while the global minima are always in the ballpark of $\sigma=0.05$.}
  \label{fig:wt-sigma-sensitivity}
\end{figure}

For \wtpgd;, we set an attack strength of $\epsilon=8/255$ and a step size of $\alpha=0.01$, as is standard practice. For \wtzoo, we set $k=100$ and $\alpha=0.01$. The number of WT samples and EoT iterations in our main experiments are both set to $m=n=16$. We justify this hyperparameter choice in the analysis of Appendix~\ref{appendix:ablation}.
Finally, selecting an appropriate value for $\sigma$ is important. If the value of $\sigma$ is too high, then the WT samples will be too far from $x$, lying on points too dissimilar to $x$ to provide an informative gradient signal. If the value of $\sigma$ is too low, the sampled points will be too close to $x$, and there will be no smoothing effect. We found that $\sigma=0.05$ is a suitable value for normalized images, and use it across all experiments. Fig.~\ref{fig:wt-sigma-sensitivity} summarises our sensitivity study on $\sigma$.

It should be mentioned that in the case of AA we do not apply EoT, as it is not a stochastic defence and therefore does not produce stochastic gradients. In addition, all stochastic models evaluated in this paper are retrained, following the instructions in the original published material, when available. As a result, the accuracy scores may not exactly reflect the scores from the original papers.

\section{Ablation Study: Selection of $m$ and $n$}
\label{appendix:ablation}

\begin{figure*}[t]
  \centering
  \begin{subfigure}{0.24\linewidth}
    \includegraphics[width=\textwidth]{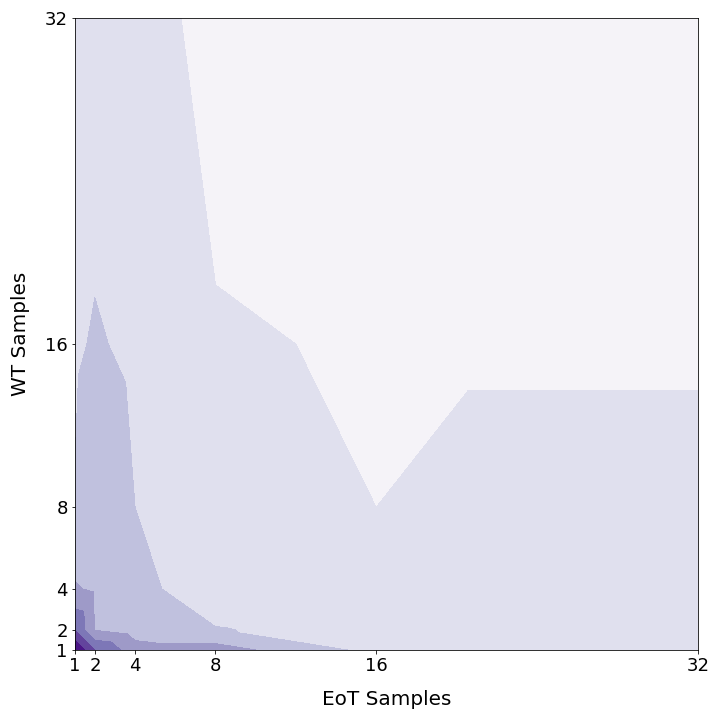}
    \caption{PNI~\cite{iccv19pni}}
    \label{fig:wt-vs-eot-pni}
  \end{subfigure}
  \hfill
  \begin{subfigure}{0.24\linewidth}
    \includegraphics[width=\textwidth]{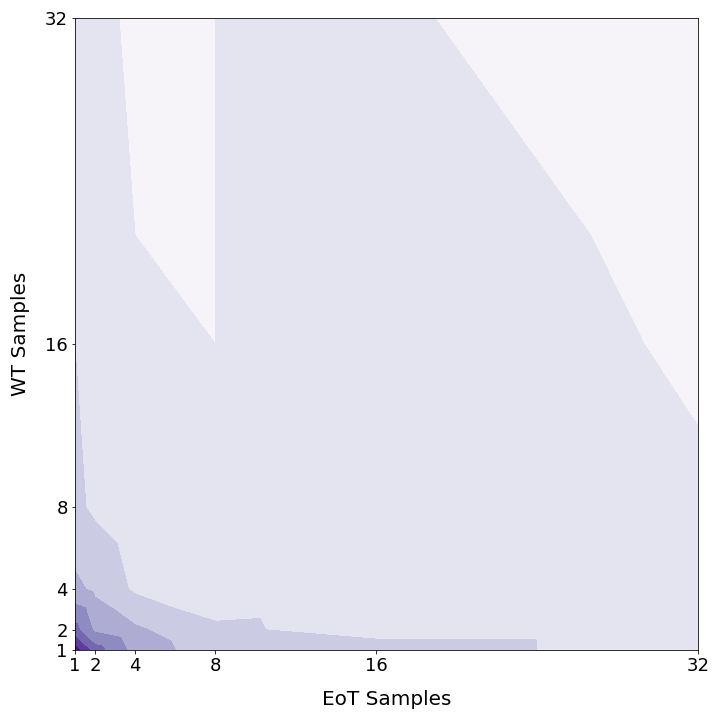}
    \caption{L2P~\cite{cvpr20l2p}}
    \label{fig:wt-vs-eot-l2p}
  \end{subfigure}
  \hfill
  \begin{subfigure}{0.24\linewidth}
    \includegraphics[width=\textwidth]{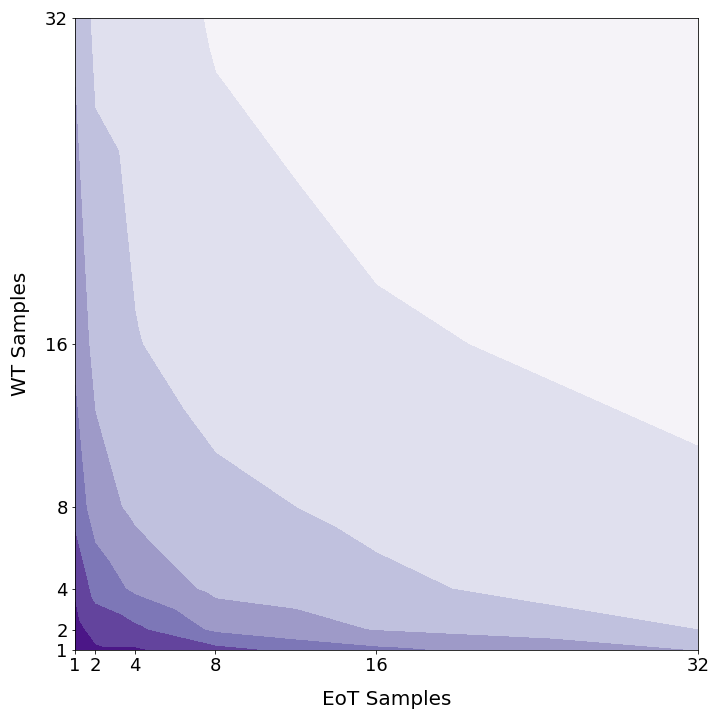}
    \caption{SE-SNN~\cite{aaai2021sesnn}}
    \label{fig:wt-vs-eot-sesnn}
  \end{subfigure}
  \hfill
  \begin{subfigure}{0.24\linewidth}
    \includegraphics[width=\textwidth]{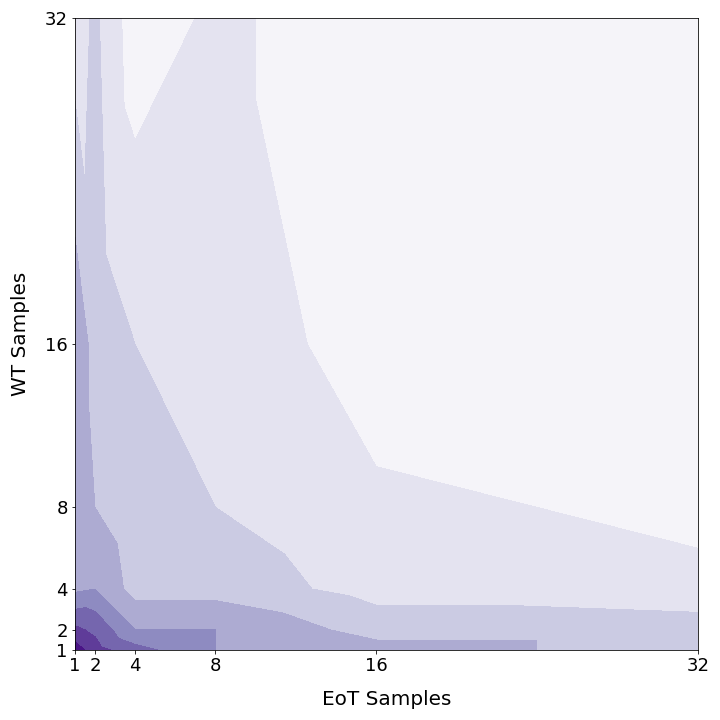}
    \caption{WCA~\cite{icml21wca}}
    \label{fig:wt-vs-eot-wca}
  \end{subfigure}
  \caption{Analysis of the interaction between WT and EoT on stochastic defences. WT and EoT are complementary. Neither can achieve peak performance alone, and best performance requires combining them (lighter color = lower accuracy). }
  \label{fig:wt-vs-eot}
\end{figure*}

We also conduct an experiment using a grid of EoT and WT samples from \{1, 2, 4, 8, 16, 32\}. 
Fig.~\ref{fig:wt-vs-eot} presents an overhead plot of the resulting network accuracy as a function of number of samples for each of EoT and WT.  Darker colors indicate higher accuracy, starting from the point (1, 1), i.e., 1 iteration of EoT and 1 WT sample (the input image itself). We see that: (i) After (16, 16) the performance of the attack quickly saturates across all defences. This justifies our use of $m=n=16$ samples in the main experiment. (ii) Even at the limit of 32 samples, neither attack method on its own performs as well as their combination. This shows that simply increasing the number of EoT samples can not replicate the effect of WT (and vice-versa).

\section{Visualising the Loss Landscapes}
\label{appendix:compare}

\begin{figure*}[t]
  \centering
  \begin{subfigure}{0.25\linewidth}
    \includegraphics[width=\textwidth]{figures/loss-landscapes-pni.png}
    \caption{PNI~\cite{iccv19pni}}
  \end{subfigure}
  \hspace*{\fill}
  \begin{subfigure}{0.25\linewidth}
    \includegraphics[width=\textwidth]{figures/loss-landscapes-l2p.png}
    \caption{L2P~\cite{cvpr20l2p}}
  \end{subfigure}
  \hspace*{\fill}
  \begin{subfigure}{0.25\linewidth}
    \includegraphics[width=\textwidth]{figures/loss-landscapes-sesnn.png}
    \caption{SE-SNN~\cite{aaai2021sesnn}}
  \end{subfigure}
  \bigskip
  \begin{subfigure}{0.25\linewidth}
    \includegraphics[width=\textwidth]{figures/loss-landscapes-smoothed-pni.png}
    \caption{PNI + \wtpgd;}
    \label{fig:loss-landscapes-smoothed-pni-2}
  \end{subfigure}
  \hspace*{\fill}
  \begin{subfigure}{0.25\linewidth}
    \includegraphics[width=\textwidth]{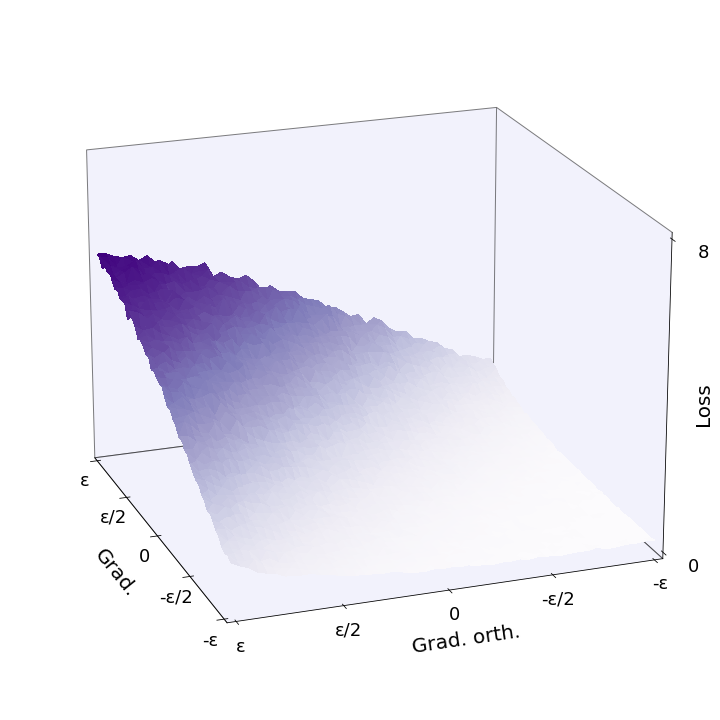}
    \caption{L2P + \wtpgd;}
    \label{fig:loss-landscapes-smoothed-l2p}
  \end{subfigure}
  \hspace*{\fill}
  \begin{subfigure}{0.25\linewidth}
    \includegraphics[width=\textwidth]{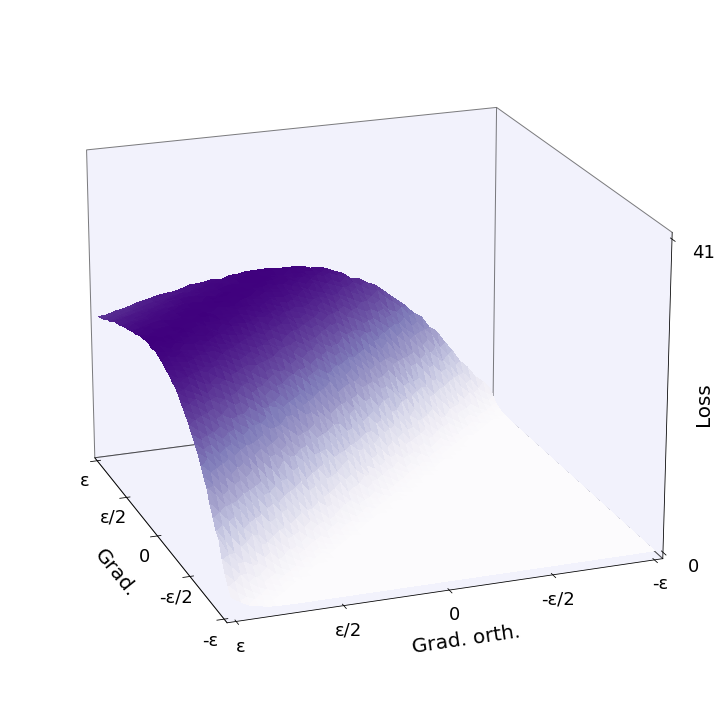}
    \caption{SE-SNN + \wtpgd;}
    \label{fig:loss-landscapes-smoothed-sesnn}
  \end{subfigure}
  \bigskip
  \begin{subfigure}{0.25\linewidth}
    \includegraphics[width=\textwidth]{figures/loss-landscapes-wca.png}
    \caption{WCA~\cite{icml21wca}}
  \end{subfigure}
  \hspace*{\fill}
  \begin{subfigure}{0.25\linewidth}
    \includegraphics[width=\textwidth]{figures/loss-landscapes-aa.png}
    \caption{AA~\cite{icml21antiadv}}
  \end{subfigure}
  \hspace*{\fill}
  \begin{subfigure}{0.25\linewidth}
    \includegraphics[width=\textwidth]{figures/loss-landscapes-kwta.png}
    \caption{k-WTA~\cite{iclr20kwta}}
  \end{subfigure}
  \bigskip
  \begin{subfigure}{0.25\linewidth}
    \includegraphics[width=\textwidth]{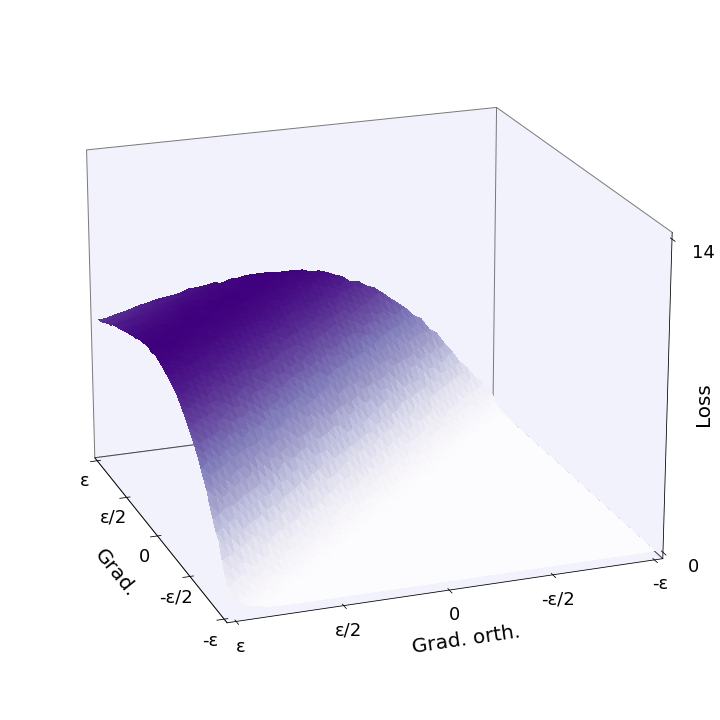}
    \caption{WCA + \wtpgd;}
    \label{fig:loss-landscapes-smoothed-wca}
  \end{subfigure}
  \hspace*{\fill}
  \begin{subfigure}{0.25\linewidth}
    \includegraphics[width=\textwidth]{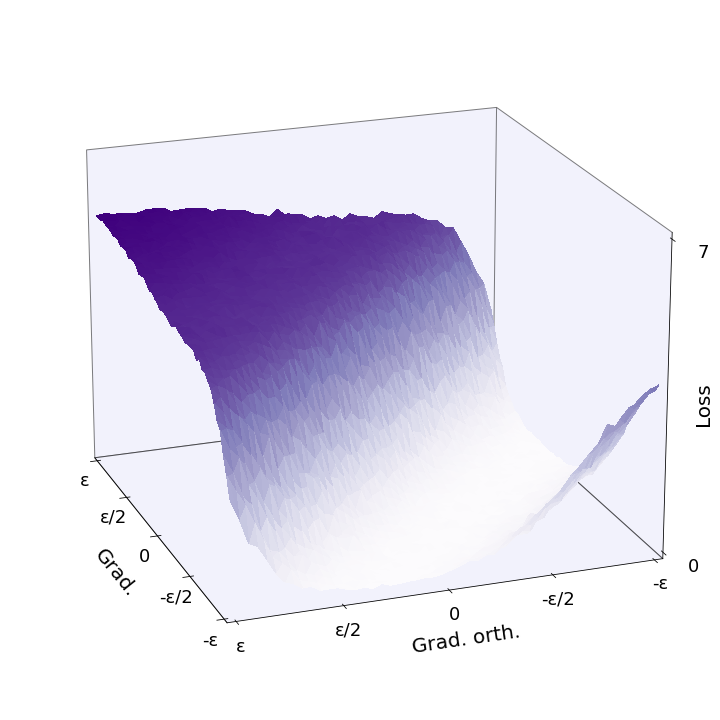}
    \caption{AA + \wtpgd;}
    \label{fig:loss-landscapes-smoothed-aa}
  \end{subfigure}
  \hspace*{\fill}
  \begin{subfigure}{0.25\linewidth}
    \includegraphics[width=\textwidth]{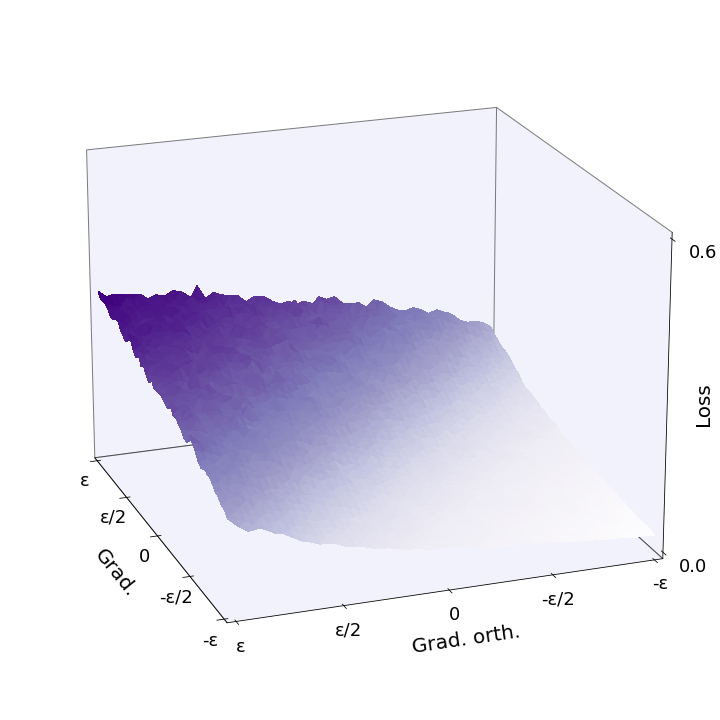}
    \caption{k-WTA + \wtpgd;}
    \label{fig:loss-landscapes-smoothed-kwta}
  \end{subfigure}
  \caption{Loss landscapes of PNI, L2P, SE-SNN, WCA, AA and k-WTA when under attack by \wtpgd;. The WT has smoothed the landscapes compared to those shown in Fig.~\ref{fig:loss-landscapes}.}
  \label{fig:loss-landscapes-smoothed}
\end{figure*}

In this Section, we describe a diagnostic method that we use to visually identify whether an adversarial defence produces a noisy loss landscape, and to generate the visualisations in Fig.~\ref{fig:loss-landscapes} and~\ref{fig:loss-landscapes-smoothed}.

Given an unperturbed input image $x$ that the target model $h_\theta$ classifies correctly as class $c$, we compute the gradient of the loss w.r.t. $x$ as $g_1 = \nabla_x\mathcal{L}(h_\theta(x, c))$. We then arbitrarily choose a dimension $g_2$, such that $g_1 \perp g_2$. Finally, we create evenly-spaced query images (and potential adversarial examples) $\tilde{x_i}$ in the $\epsilon$-ball of $x$ as
\begin{equation}
  \tilde{x_i} = x + \epsilon_1 \sign(g_1) +  \epsilon_2 \sign(g_2)\;,
\end{equation}
where $\epsilon_1,\epsilon_2\in[-\frac{8}{255},\frac{8}{255}]$, and project their calculated loss values $\mathcal{L}(h_\theta(\tilde{x_i}, c))$ to the $g_1$ and $g_2$ axes.

Fig.~\ref{fig:loss-landscapes} shows the above 2D slice through the loss landscapes of PNI, L2P, SE-SNN, WCA, AA and k-WTA defences. In Fig.~\ref{fig:loss-landscapes-smoothed} we show the corresponding smoothed loss landscapes, when under attack by \wtpgd;, side-by-side for easier means of visual comparison.
Further, Appendix~\ref{appendix:non-obfuscating} includes the loss surfaces of the highest scoring non-stochastic adversarial defences listed in RobustBench~\cite{neurips21robustbench}, to give the reader a frame of reference of how non-rugged loss landscapes should look like in state-of-the-art defences.

\section{Strong Defences with Smooth Loss Landscapes}
\label{appendix:non-obfuscating}

In the main paper, we see the effect of our attack on gradient-obfuscating adversarial defences that construct a noisy loss landscape to confuse the adversary. To further support future adversarial defence research, in this Section we want to inform the reader about how the loss landscapes of non-obfuscating defences should look like.

\begin{figure*}[t]
  \centering
  \begin{subfigure}{0.32\linewidth}
    \includegraphics[width=\textwidth]{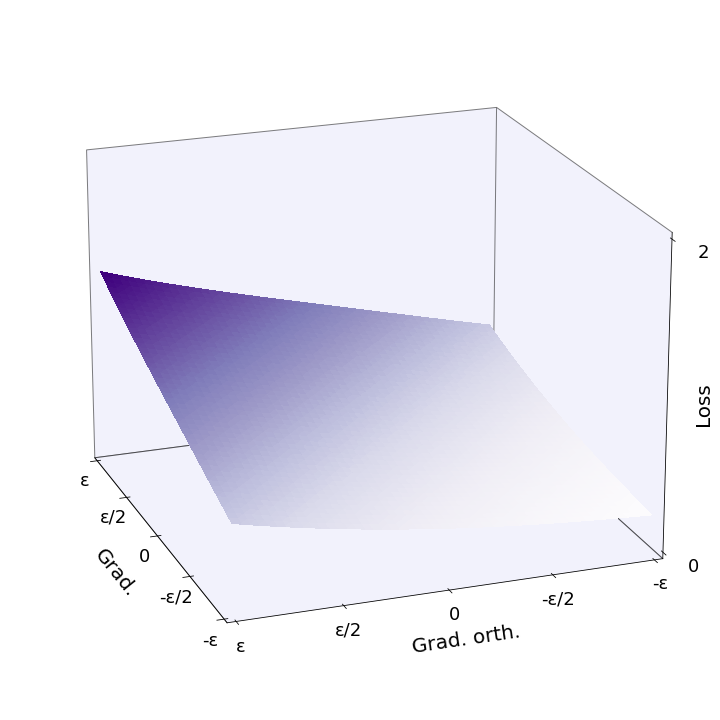}
    \caption{Rebuffi et al.~\cite{arxiv21rebuffi}}
  \end{subfigure}
  \hspace*{\fill}
  \begin{subfigure}{0.32\linewidth}
    \includegraphics[width=\textwidth]{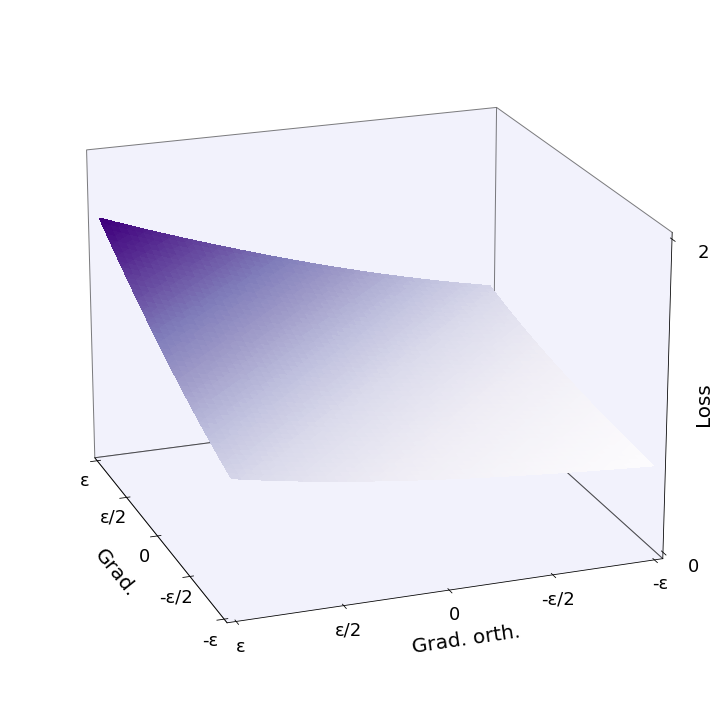}
    \caption{Gowal et al.~\cite{arxiv21gowal}}
  \end{subfigure}
  \hspace*{\fill}
  \begin{subfigure}{0.32\linewidth}
    \includegraphics[width=\textwidth]{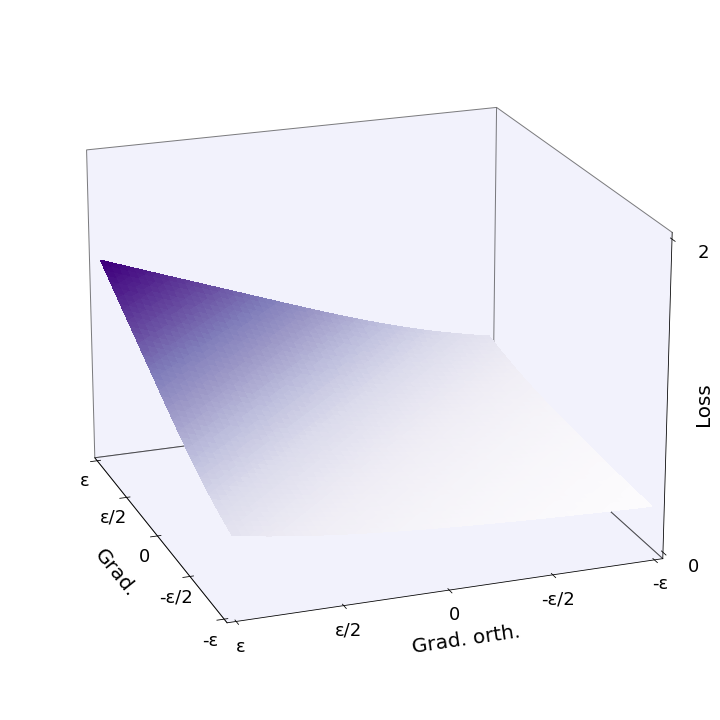}
    \caption{Rade et al.~\cite{icml21rade}}
  \end{subfigure}
  \bigskip
  \begin{subfigure}{0.32\linewidth}
    \includegraphics[width=\textwidth]{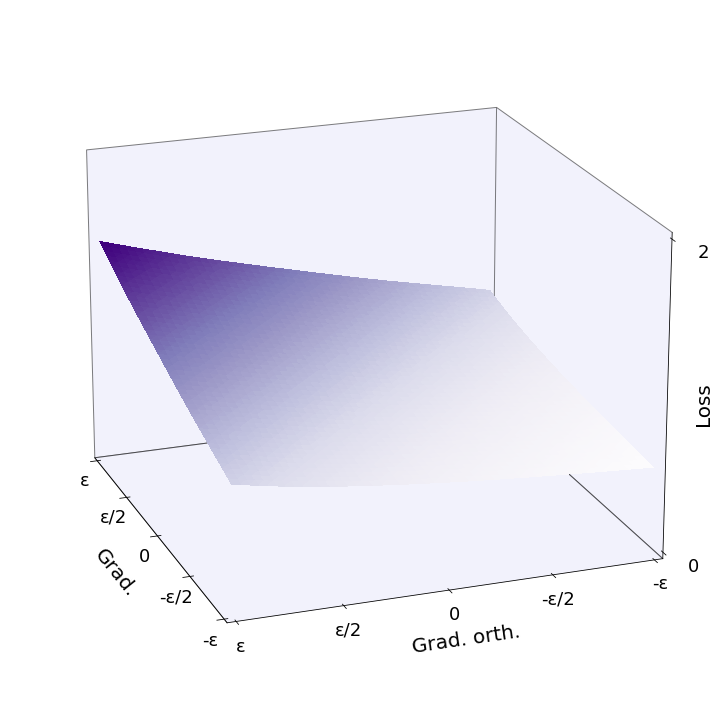}
    \caption{Sridhar et al.~\cite{arxiv2021sridhar}}
  \end{subfigure}
  \hspace*{\fill}
  \begin{subfigure}{0.32\linewidth}
    \includegraphics[width=\textwidth]{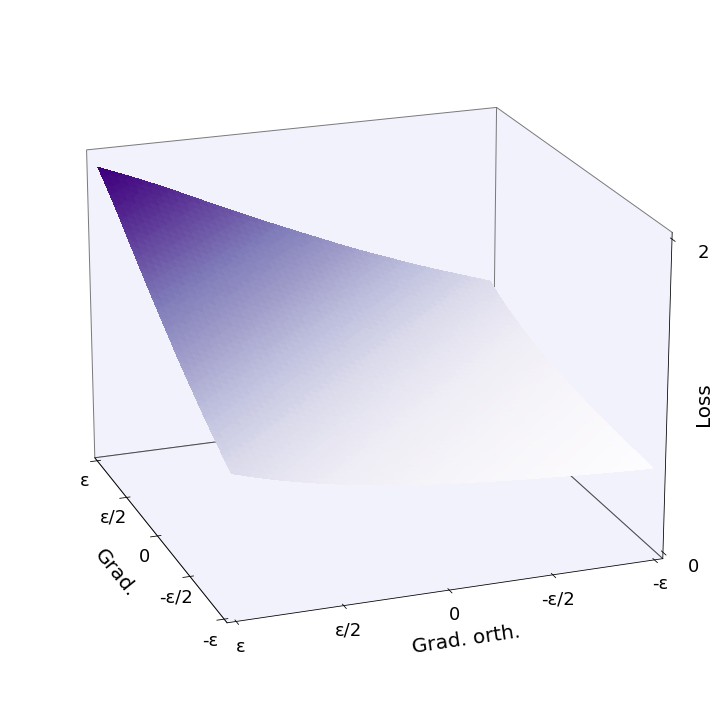}
    \caption{Wu et al.~\cite{neurips20wu}}
  \end{subfigure}
  \hspace*{\fill}
  \begin{subfigure}{0.32\linewidth}
    \includegraphics[width=\textwidth]{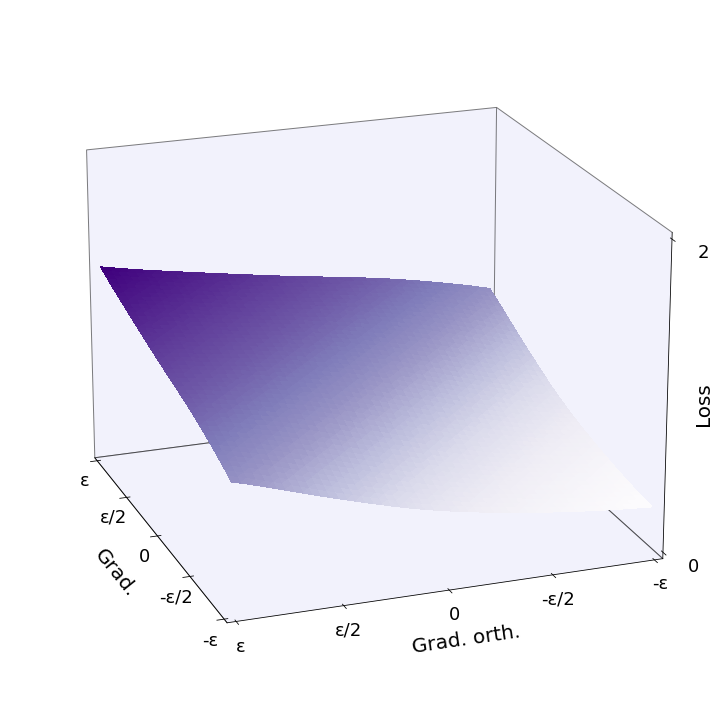}
    \caption{Zhang et al.~\cite{iclr21zhang}}
  \end{subfigure}
  \bigskip
  \begin{subfigure}{0.32\linewidth}
    \includegraphics[width=\textwidth]{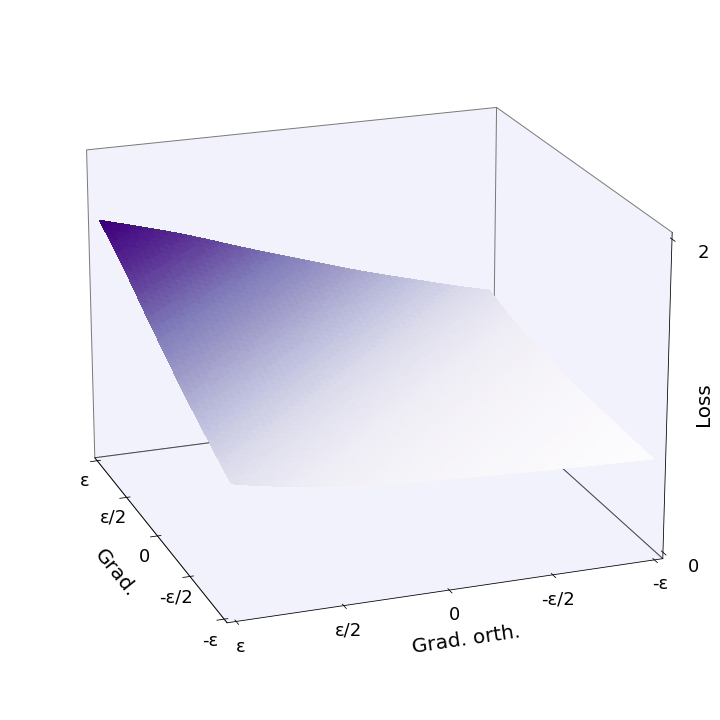}
    \caption{Carmon et al.~\cite{neurips19carmon}}
  \end{subfigure}
  \hspace*{\fill}
  \begin{subfigure}{0.32\linewidth}
    \includegraphics[width=\textwidth]{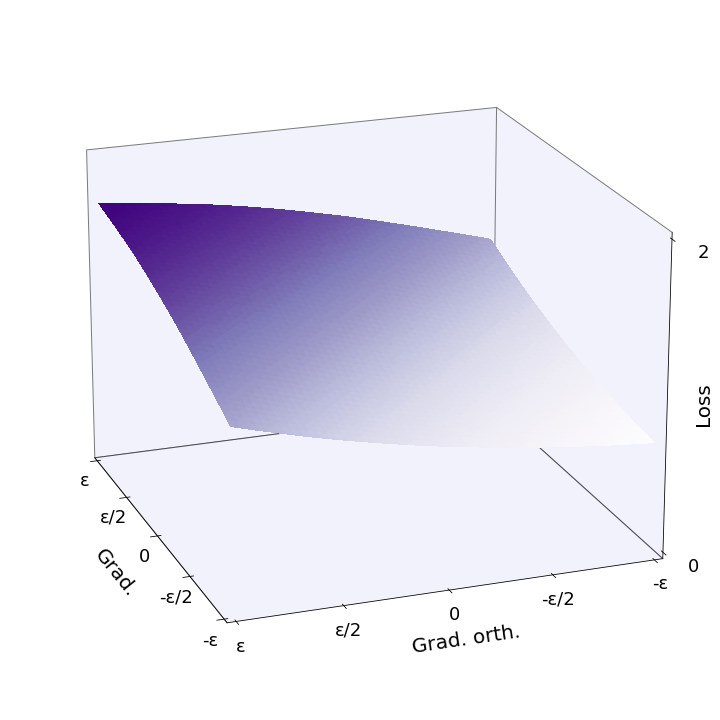}
    \caption{Wang et al.~\cite{iclr20wang}}
  \end{subfigure}
  \hspace*{\fill}
  \begin{subfigure}{0.32\linewidth}
    \includegraphics[width=\textwidth]{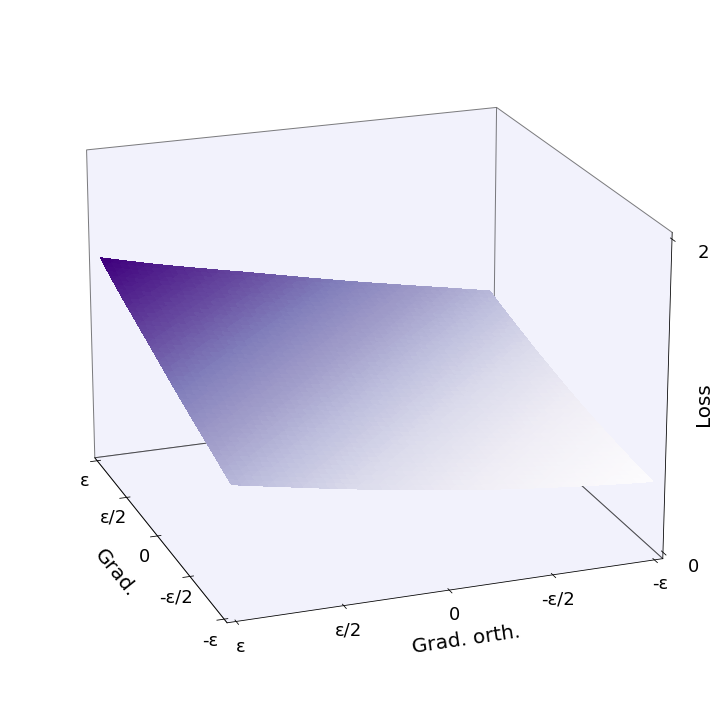}
    \caption{Hendrycks et al.~\cite{icml19hendrycks}}
  \end{subfigure}
  \hspace*{\fill}
  \caption{Landscapes of non-obfuscating adversarial defences that score competitively on RobustBench~\cite{neurips21robustbench}.}
  \label{fig:supp-smooth-surfaces}
\end{figure*}

To that end, we choose the 9 highest-scoring adversarial defences from the $\ell_\infty$ CIFAR-10 leaderboard of the widely used RobustBench~\cite{neurips21robustbench} and visualise their loss landscapes in Fig.~\ref{fig:supp-smooth-surfaces}. The visualisation method is the same that produced Fig. 1 of the main paper; except that none of the defences are stochastic and therefore EoT is not used to obtain better gradient estimates.

\newpage

\end{document}